\documentclass[letterpaper]{article}
\usepackage{aaai18}
\usepackage{times}
\usepackage{helvet}
\usepackage{courier}
\usepackage{url}
\usepackage{graphicx}
\usepackage{booktabs}
\usepackage{subfigure}
\usepackage{tabularx}
\usepackage{multirow}
\usepackage{wrapfig}
\usepackage{amssymb}
\usepackage{color}
\graphicspath{{figures/}}
\DeclareGraphicsExtensions{.pdf}

\title{Mix-and-Match Tuning for Self-Supervised Semantic Segmentation}
\author{Xiaohang Zhan \quad Ziwei Liu \quad Ping Luo \quad Xiaoou Tang \quad Chen Change Loy \\
Department of Information Engineering, The Chinese University of Hong Kong \\
\{zx017, lz013, pluo, xtang, ccloy\}@ie.cuhk.edu.hk}
\begin{document}
\maketitle


\begin{abstract}

Deep convolutional networks for semantic image segmentation typically require large-scale labeled data, e.g.,~ImageNet and MS COCO, for network pre-training.
To reduce annotation efforts, self-supervised semantic segmentation is recently proposed to pre-train a network without any human-provided labels.
The key of this new form of learning is to design a proxy task (e.g.,~image colorization), from which a discriminative loss can be formulated on unlabeled data.
Many proxy tasks, however, lack the critical supervision signals that could induce discriminative representation for the target image segmentation task. Thus self-supervision's performance is still far from that of supervised pre-training.
In this study, we overcome this limitation by incorporating a `mix-and-match' (M\&M) tuning stage in the self-supervision pipeline. The proposed approach is readily pluggable to many self-supervision methods and does not use more annotated samples than the original process. Yet, it is capable of boosting the performance of target image segmentation task to surpass fully-supervised pre-trained counterpart.
The improvement is made possible by better harnessing the limited pixel-wise annotations in the target dataset. Specifically, we first introduce the `mix' stage, which sparsely samples and mixes patches from the target set to reflect rich and diverse local patch statistics of target images. A `match' stage then forms a class-wise connected graph, which can be used to derive a strong triplet-based discriminative loss for fine-tuning the network.
Our paradigm follows the standard practice in existing self-supervised studies and no extra data or label is required.
With the proposed M\&M approach, for the first time, a self-supervision method can achieve comparable or even better performance compared to its ImageNet pre-trained counterpart on both PASCAL VOC2012 dataset and CityScapes dataset.

\end{abstract}

\section{Introduction}

Semantic image segmentation is a classic computer vision task that aims at assigning each pixel in an image with a class label such as ``chair'', ``person'', and ``dog''.
It enjoys a wide spectrum of applications, such as scene understanding~\cite{li2009towards,lin2014microsoft,li2017video} and autonomous driving~\cite{geiger2013vision,cordts2016cityscapes,li2017not}.
Deep convolutional neural network (CNN) is now the state-of-the-art technique for semantic image segmentation~\cite{long2015fully,liu2015semantic,zhao2016pyramid,liu2017deep}. The excellent performance, however, comes with a price of expensive and laborious label annotations.
In most existing pipelines, a network is usually first pre-trained on millions of class-labeled images, e.g.,~ImageNet \cite{russakovsky2015imagenet} and MS COCO \cite{lin2014microsoft}, and subsequently fine-tuned with thousands of pixel-wise annotated images.


Self-supervised learning\footnote{Project page: http://mmlab.ie.cuhk.edu.hk/projects/M\&M/} is a new paradigm proposed for learning deep representations without extensive annotations. This new technique has been applied to the task of image segmentation~\cite{zhang2016colorful,larsson2016learning,larsson2017colorization}.
In general, self-supervised image segmentation can be divided into two stages: \textit{the proxy stage}, and \textit{the fine-tuning stage}.
The proxy stage does not need any labeled data but requires one to design a proxy or pretext task with self-derived supervisory signals on unlabeled data.
For instance, learning by colorization~\cite{larsson2017colorization} utilizes the fact that a natural image is composed of luminance channel and chrominance channels. The proxy task is formulated with cross-entropy loss to predict an image chrominance from the luminance of the same image.
In the fine-tuning stage, the learned representations are utilized to initialize the target semantic segmentation network. The network is then fine-tuned with pixel-wise annotations.
It has been shown that without large-scale class-labeled pre-training, semantic image segmentation could still gain encouraging performance over random initialization or from-scratch training. 


Though promising, the performance of self-supervised learning is still far from that achieved by supervised pre-training. 
For instance, a VGG-16 network trained with the self-supervised method of~\cite{larsson2017colorization} achieves a 56.0\% mean Intersection over Union (mIoU) on PASCAL VOC 2012 segmentation benchmark~\cite{everingham2010pascal}, higher than a random initialized network that only yields 35.0\% mIoU. However, an identical network trained on ImageNet achieves 64.2\% mIoU. There exists a considerable gap between self-supervised and pure supervised pre-training.

\begin{figure}[t]
\centering
\includegraphics[width=\linewidth]{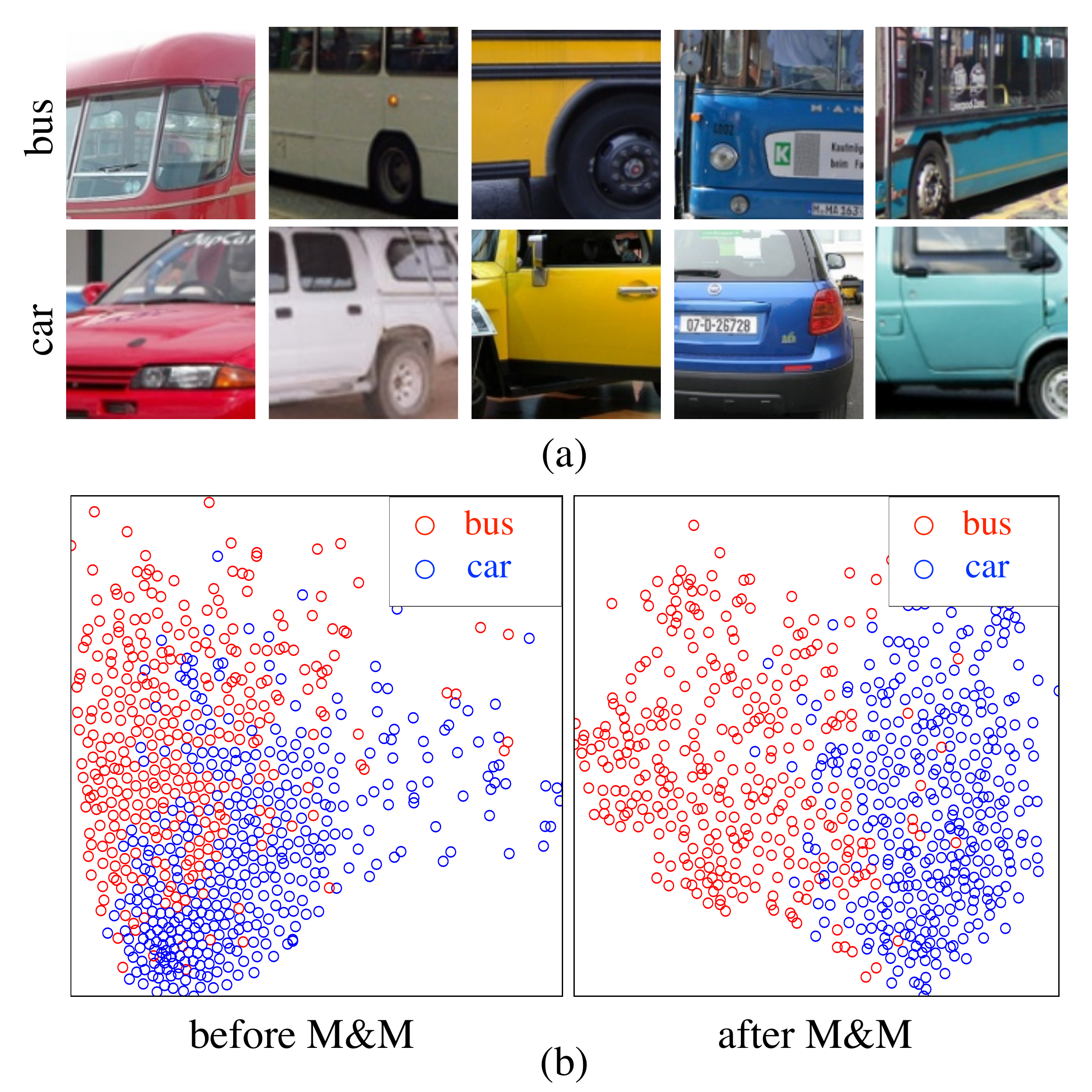}
\caption{\small{(a) shows samples of patches from categories `bus' and `car', and these two categories have similar color distributions but different patch statistics. (b) depicts deep feature distributions of `bus' and `car', before and after mix-and-match, visualized with t-SNE~\cite{maaten2008visualizing}. Best viewed in color.}}
\label{fig:intro}
\end{figure}

We believe that the performance discrepancy is mainly caused by the semantic gap between the proxy task and the target task. Take learning by colorization as an example, the goal of the proxy task is to colorize gray-scale images. 
The representations learned from colorization may be well-suited for modeling color distributions, but are likely amateur in discriminating high-level semantics.
For instance, as shown in Fig.~\ref{fig:intro}(a), a red car can be arbitrarily more similar to a red bus than to a blue car. The features of both car and bus classes are highly overlapped, as depicted by the feature embedding in the left plot of Fig.~\ref{fig:intro}(b). 


Improving the performance of self-supervised image segmentation requires one to improve the discriminative power of representation tailored to the target task. This goal is non-trivial -- target's pixel-wise annotations are discriminative for the goal but they often available with just a handful amount, typically in thousands of labeled images. 
Existing approaches typically use a pixel-wise softmax loss to exploit pixel-wise annotations for fine-tuning a network.
This strategy may be sufficient for a network that is well-initialized by supervised pre-training but could fall inadequate for a self-supervised network of which the features are weak.
We argue that pixel-wise softmax loss is not the sole way of harnessing the information provided by pixel-wise annotations.


In this study, we present a new learning strategy called `mix-and-match' (M\&M), which can help harness the scarce labeled information of a target set for improving the performance of networks pre-trained by self-supervised learning. The M\&M learning is conducted after the proxy stage and before the usual target fine-tuning stage, serving as an intermediate step to bridge the gap between the proxy and target tasks. It is noteworthy that M\&M only uses the target images and its labels thus no additional annotation is required. 

The essence of M\&M is inspired by metric learning. 
In the `\textit{mix}' step, we randomly sample a large number of local patches from the target set and mix them together. 
The patch set is formed across images thus decouple any intra-image dependency to faithfully reflect the diverse and rich target distribution. Extracting patches also allows us to generate a massive number of triplets from the small target image set to produce stable gradients for training our network.
In the `\textit{match}' step, we form a graph with nodes defined by patches represented by their deep features. An edge between nodes is defined as attractive if the nodes share the same class label; otherwise, it is a rejective edge.
We enforce a class-wise connected graph, that is, all nodes from the same class in the graph compose a connected subgraph, as shown in Fig.~\ref{fig:graph}(c). This ensures global consistency in triplet selection coherent to the class labels. With the graph, we can derive a robust triplet loss that encourages the network to map each patch to a point in feature space so that patches belonging to the same class lie close together while patches of different classes are separated by a wide margin. 
The way we sample triplets from a class-wise connected graph differs significantly from existing approach~\cite{schroff2015facenet} that forms multiple disconnected subgraphs for each class.
%


We summarize our contributions as follows. 1) We formulate a novel `mix-and-match' tuning method, which for the first time, allows networks pre-trained with self-supervised learning to outperform the supervised learning counterpart. Specifically, with VGG-16 as the backbone network, by using image colorization as the proxy task, our M\&M method achieves 64.5\%, outperforming the ImageNet pre-trained network that achieves 64.2\% mIoU on PASCAL VOC2012 dataset. Our method also obtains 66.4\% mIoU on CityScapes dataset, comparable to 67.9\% mIoU achieved by using a ImageNet pre-trained network. This improvement is significant considering that our approach is based on unsupervised pre-training.
2) Apart from the learning by colorization method, M\&M also improves learning by context method~\cite{noroozi2016unsupervised} by a large margin.
3) In the setting of random initialization, our method achieves significant improvements with both AlexNet and VGG-16, on both PASCAL VOC2012 and CityScapes. It makes training semantic segmentation from scratch possible.
4) In addition to the new notion of mix-and-match, we also present a triplet selection mechanism based on class-wise connected graph, which is more robust than conventional selection scheme for our task.


\section{Related Work}
\label{sec:relatedwork}

\noindent
\textbf{Self-supervision.}
It is a standard and established practice to pre-train a deep network with large-scale class-labeled images (e.g., ImageNet) before fine-tuning the model for other visual tasks. Recent research efforts are gearing towards reducing the degree of or eliminating supervised pre-training altogether. Among various alternatives, self-supervised learning is gaining substantial interest. 
To enable self-supervised learning, proxy tasks are designed so that meaningful representations can be induced from the problem-solving process.
Popular proxy tasks include sample reconstruction~\cite{pathak2016context}, temporal correlation~\cite{wang2015unsupervised,pathak2016learning}, learning by context~\cite{doersch2015unsupervised,noroozi2016unsupervised}, cross-transform correlation~\cite{dosovitskiy2015discriminative} and learning by colorization~\cite{zhang2016colorful,zhang2016split,larsson2016learning,larsson2017colorization}.
In this study, we do not design a new proxy task, but present an approach that could uplift the discriminative power of a self-supervised network tailored to the image segmentation task. We demonstrate the effectiveness of M\&M on learning by colorization and learning-by-context.

\noindent
\textbf{Weakly-supervised segmentation.}
There exists a rich body of literature that investigates approaches for reducing annotations in learning deep models for the task of image segmentation.
Alternative annotations such as point~\cite{bearman2016s}, bounding box~\cite{dai2015boxsup}~\cite{papandreou2015weakly}, scribble~\cite{lin2016scribblesup} and video~\cite{hong2017weakly} have been explored as ``cheap'' supervisions to replace the pixel-wise counterpart.
Note that these methods still require ImageNet classification as a pre-training task.
Self-supervised learning is more challenging in that no image-level supervision is provided in the pre-training stage. The proposed M\&M approach is dedicated to improve the weak representation learned by self-supervised pre-training.

\noindent
\textbf{Graph-based segmentation.}
Graph-based image segmentation~\cite{felzenszwalb2004efficient} has been explored from early years. The main idea is to explore dependency between pixels.
Different from the conventional graph on pixels or superpixels in a single image, the proposed method defines the graph on image patches sampled from multiple images. We do not partition image by performing cuts on a graph, but use the graph to select triplets for the proposed discriminative loss.

\section{Mix-and-Match Tuning}

Figure~\ref{fig:arch} illustrates the proposed approach, where (a) and (c) depict the conventional stages for self-supervised semantic image segmentation, while (b) shows the proposed `mix-and-match' (M\&M) tuning.
Specifically, in (a), a proxy task, e.g.,~learning by colorization, is designed to pre-train the CNN using unlabeled images.
In (c), the pre-trained CNN is fine-tuned on images and the associated per-pixel labeled maps of a target task.
This work inserts M\&M tuning between the proxy task and the target task as shown in (b). It is noteworthy that M\&M uses the same target images and label maps in (c), hence no additional data is required.
As the name implies, M\&M tuning consists of two steps, namely `mix' and `match'. We explain these steps as follows.

\subsection{The Mix Step -- Patch Sampling}
Recall that our goal is to better harness the information in pixel-wise annotations of the target set.
Image patches have long been considered as strong visual primitive~\cite{singh2012unsupervised} that incorporates both appearance and structure information. Visual patches have been successfully applied to various tasks in visual understanding~\cite{li2013harvesting}. Inspired by these pioneering works, the first step of M\&M tuning is designed to be a `mix' step that aims at sampling patches across images. The relation between these patches can be exploited for optimization in the subsequent `match' operation.

\begin{figure}[t]
\centering
\includegraphics[width=\linewidth]{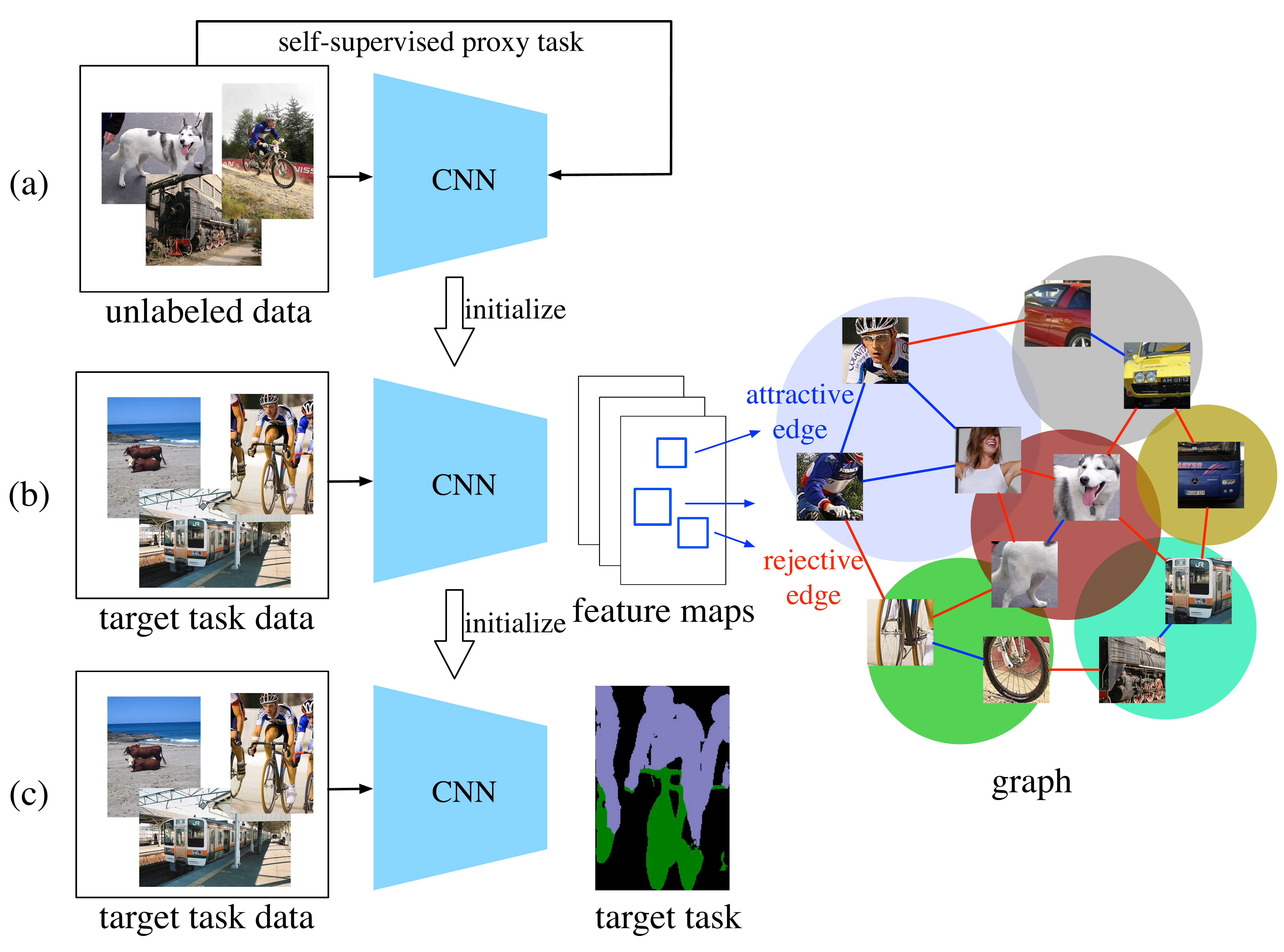}
\caption{\small{An overview of the mix-and-match approach. Our approach starts with a self-supervised proxy task (a), and uses the learned CNN parameters to initialize the CNN in mix-and-match tuning (b). Given an image batch with label maps of the target task, we select and mix image patches and then match them according to their classes via a class-wise connected graph. The matching gives rise to a triplet loss, which can be optimized to tune the parameters of the network via back propagation. Finally, the modified CNN parameters are further fine-tuned to the target task (c).}}
\label{fig:arch}
\end{figure}

More precisely, a large number of image patches with various spatial sizes are randomly sampled from a batch of images. Heavily overlapped patches are discarded. These patches are represented by using the features extracted from the CNN pre-trained in the stage of Fig.~\ref{fig:arch}(a), and assigned with unique class labels based on the corresponding label map.
The patches across all images are mixed to decouple any intra-image dependency so as to reflect the diverse and rich target distribution.
The mixed patches are subsequently utilized as the input for the `match' operation.

\subsection{The Match Step -- Perceptual Patch Graph}

Our next goal is to exploit the patches to generate stable gradients for tuning the network.
This is possible since patches are of different classes, and such relation can be employed to form a massive number of triplets. A triplet is denoted as $(P_a, P_p, P_n)$, where $P_a$ is an anchor patch, $P_p$ is a positive patch that shares the same label as $P_a$, and $P_n$ is a negative patch with a different class label.
With the triplets, one can formulate a discriminative triplet loss for fine-tuning the network.

\begin{figure}[t]
\centering
\includegraphics[width=\linewidth]{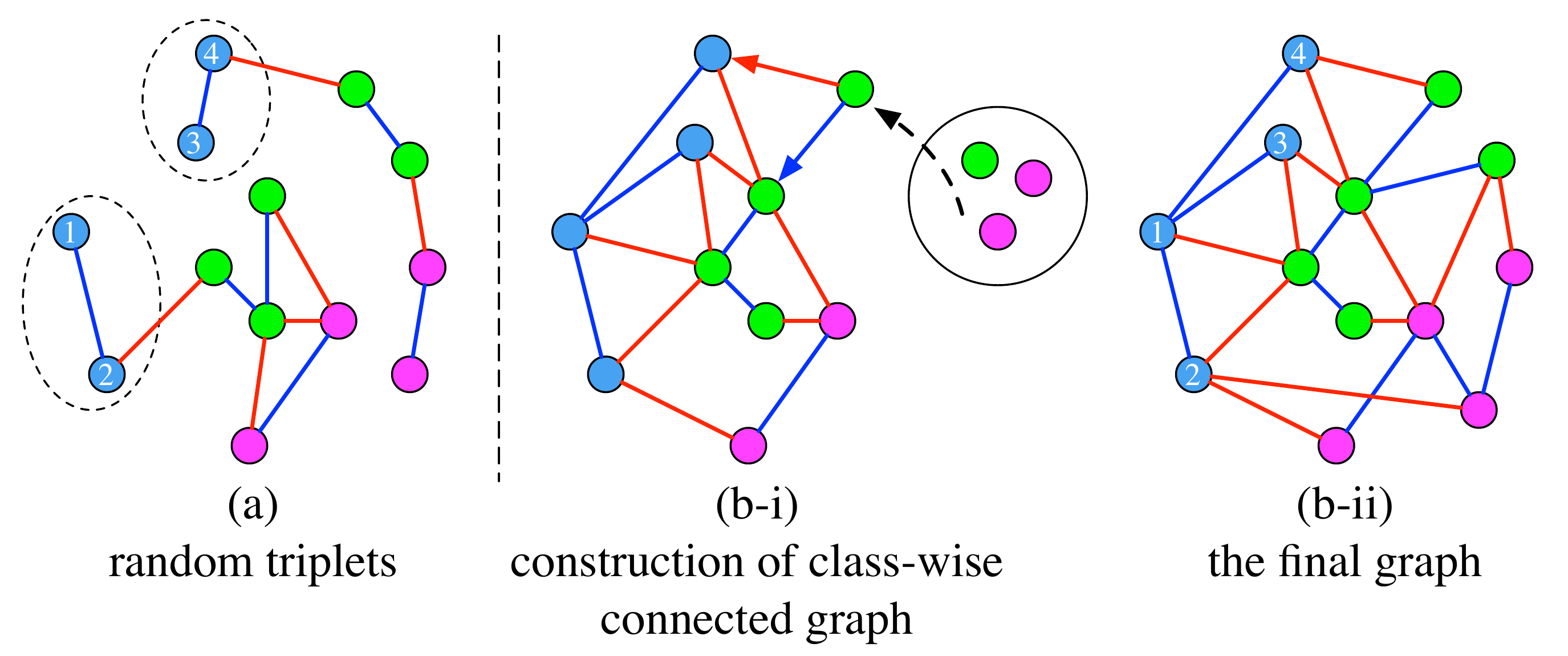}
\caption{\small{This figure shows different strategies of drawing triplets. The color of nodes represent their labels. Blue and red edges denote attractive and rejective edges, respectively. (a) depicts the random triplet strategy~\cite{schroff2015facenet}, where nodes from the same class do not necessarily form a connected subgraph. (b-i) and (b-ii) shows the proposed triplet selection strategy. A class-wised connected graph is constructed to sample triplets, which enforces tighter constraints on positive class boundary. Details are explained in the main text of methodology section. Best viewed in color.}}
\label{fig:graph}
\end{figure}

A conventional way of sampling triplets is to follow the notion of~\citeauthor{schroff2015facenet}~\shortcite{schroff2015facenet}. For convenience, we call this strategy as `\textit{random triplets}'. In this strategy, triplets are randomly picked from the input batch.
For instance, as shown in Fig.~\ref{fig:graph}(a), nodes $\{1,2\}$ and an arbitrary negative patch forms a triplet, and nodes $\{3,4\}$ and another negative patch forms another triplet.
As can be seen, there is no positive connection between nodes $\{1,2\}$ and $\{3,4\}$ despite they share a common class label.
While locally the distance between each triplet is optimized, the boundary of the positive class can be loose since the global constraint (i.e.~all nodes $\{1,2,3,4\}$ must lie closer) is not enforced.
We term this phenomenon as global inconsistency.
Empirically, we found that this approach tends to perform poorer than the proposed method, which will be introduced next.

The proposed `match' step draws triplets in a different way from the conventional approach~\cite{schroff2015facenet}. 
In particular, the `match' step begins with graph construction based on the mixed patches. For each CNN learning iteration, we construct a graph on-the-fly given a batch of input images. The nodes of the graph are patches.
Two types of edges are defined between nodes -- a) ``attractive'' if two nodes have an identical class label and b) ``rejective'' if two nodes have different class labels.
Different from~\cite{schroff2015facenet}, we enforce the graph to be connected, and importantly, the graph should be class-wise connected. That is, all nodes from the same class in the graph compose a connected subgraph via ``attractive'' edges.
We adopt an iterative strategy to create such a graph.
At first, the graph is initialized to be empty. Then, as shown in Fig.~\ref{fig:graph}(b), patches are absorbed individually into the graph as a node and it creates respectively one ``attractive'' and ``rejective'' edge with existing nodes in the graph.

An example of an established graph is shown in Fig.~\ref{fig:graph}(c).
Considering nodes $\{1,2,3,4\}$ again, unlike `random triplets', the nodes form a connected subgraph. Different classes represented in green nodes and pink nodes also form coherent clusters based on their respective classes, imposing tighter constraints than random triplets.
To fully realize such class-wise constraints, each node in the graph will take turn to serve as an anchor for loss optimization.
An added benefit of permitting all nodes as possible anchor candidate is the improved utilization efficiency of patch relation over random triplets.

\subsection{The Tuning Loss}
\label{subsec:train}
\noindent
\textbf{Loss function.}
To optimize the semantic consistency within the graph, for any two nodes in the graph, if they are connected by attractive edges, we seek to minimize their distance in the feature space; and if they are connected by rejective edges, the distance should be maximized.
Consider a node that connects two other nodes via attractive and rejective edges, we denote it as an ``anchor'' while the two connected nodes are denoted as ``positive'' and ``negative'' respectively.
These three nodes are grouped to be a ``triplet''.

When constructing the graph, we ensure that each node can serve as ``anchor'', except for those nodes whose labels are unique among all the nodes.
Thus, the number of nodes equals the number of triplets.
Assume that in each iteration we discover $N$ triplets in the graph.
By converting the graph optimization problem into ``triplet ranking''~\cite{schroff2015facenet}, we formulate our loss function as follows:
\begin{equation}
L = \frac{1}{N}\sum_{i}^{N}\max\left \{ D\left ( P_a^i, P_p^i \right ) - D\left ( P_a^i, P_n^i \right ) + \alpha, 0 \right \},
\end{equation}
where $P_a$, $P_p$, $P_n$, denote ``anchor'', ``positive'', ``negative'' nodes in a triplet, 
$\alpha$ is a regularization factor controlling the distance margin and $D(\cdot,\cdot)$ is a distance metric measuring patch relationship.

In this work, we leverage perceptual distance~\cite{gatys2015neural} to characterize the relationship between patches.
This is different from previous works~\cite{singh2012unsupervised}~\cite{doersch2012makes} that define patch distance using low-level cues (e.g.,~colors and edges). 
Specifically, the perceptual representation can be formulated as $f: P\rightarrow \mathbf{x}$, where $f$ denotes a convolutional neural network (CNN) and $\mathbf{x}$ denotes the extracted representation.
$D(P_0, P_1)$ is the perceptual distance between two patches, which can be formulated as:
\begin{equation}
D(P_i, P_j) = \left \| \left( \mathbf{x}_i / \|\mathbf{x}_i\|_2 - \mathbf{x}_j / \|\mathbf{x}_j\|_2 \right) \right \| ^{2},
\end{equation}
where $\mathbf{x}_i$ and $\mathbf{x}_j$ is the CNN representation extracted from patch $P_i$ and $P_j$.
$L_{2}$ normalization is used here for calculating Euclidean distances.

By optimizing the ``triplet ranking'' loss, our perceptual patch graph converges to both intra-class and inter-class semantic consistency.

\noindent
\textbf{M\&M implementation details.}
We use both AlexNet~\cite{krizhevsky2012imagenet} and VGG-16~\cite{simonyan2015very} as our backbone CNN architectures, as illustrated in Fig.~\ref{fig:arch}.
For initialization, we try random initialization and two proxy tasks including Jigsaw Puzzles~\cite{noroozi2016unsupervised} and Colorization~\cite{larsson2017colorization}.
From a batch of $16$ images in each CNN iteration, we sample $10$ patches per image with various sizes and resize them to a fixed size of $128\times128$.
Then we extract ``pool5'' feature of these patches from the CNN for later usage.
We assign the patches' unique labels as the central pixel labels using the corresponding label maps.
Then we perform the iterative strategy to construct the graph as discussed in the methodology section.
We make use of each node in the graph as an ``anchor'', which is made possible by our graph construction strategy. If any node whose label is unique among all the nodes, we duplicate it as its ``positive'' counterpart.
In this way, we obtain a batch of meaningful triplets whose number is equal to the number of nodes, and feed them into a triplet loss layer, whose margin $\alpha$ is set as $2.1$.
Such a M\&M tuning is conducted for 8000 iterations on PASCAL VOC2012 or CityScapes training dataset.
The learning rate is fixed at $0.01$ before iteration 6000, and then dropped to $0.001$. We apply batch normalization to speed up convergence.

\noindent
\textbf{Segmentation fine-tuning details.}
Finally, we fine-tune the CNN to the semantic segmentation task. For AlexNet, we follow the same setting as presented in~\cite{noroozi2016unsupervised}, and for VGG-16, we follow~\cite{larsson2017colorization} whose architecture is equipped with hyper-columns~\cite{hariharan2015hypercolumns}.
The fine-tuning process undergoes 40k iterations, with an initial learning rate as 0.01 and dropped with a factor of 10 at iteration 24k, 36k.
We keep tuning batch normalization layers before ``pool5''. All experiments follow the same setting.

\section{Experiments}
\label{sec:experiments}

\noindent
\textbf{Settings.}
Different proxy tasks are combined with our M\&M tuning to demonstrate its merits.
In our experiments, as initialization, we use released models of different proxy tasks from learning by context (or Jigsaw Puzzles)~\cite{noroozi2016unsupervised} and learning by colorization~\cite{larsson2017colorization}. Both methods adopt 1.3 million unlabeled images in ImageNet dataset~\cite{deng2009imagenet} for training.
Besides that, we also perform experiments on randomly initialized networks.
In M\&M tuning, we make use of PASCAL VOC2012 dataset~\cite{everingham2010pascal}, which consists of 10,582 training samples with pixel-wise annotations. The same dataset is used in ~\cite{noroozi2016unsupervised,larsson2017colorization} for fine-tuning so no additional data is used in M\&M.
For fair comparisons, all self-supervision methods are benchmarked on PASCAL VOC2012 validation set that comes with 1,449 images.
We show the benefits of M\&M tuning on different backbone networks, including AlexNet and VGG-16.
To demonstrate the generalization ability of our learned model, we also report the performance of our VGG-16 full model on PASCAL VOC2012 test set.
We further apply our method on the CityScapes dataset~\cite{cordts2016cityscapes}, with 2,974 training samples and report results on the 500 validation samples.
All results are reported in mean Intersection over Union (mIoU), which is the standard evaluation criterion of semantic segmentation.

\subsection{Results}

\begin{figure}[t]
    \centering
    \includegraphics[width=\linewidth]{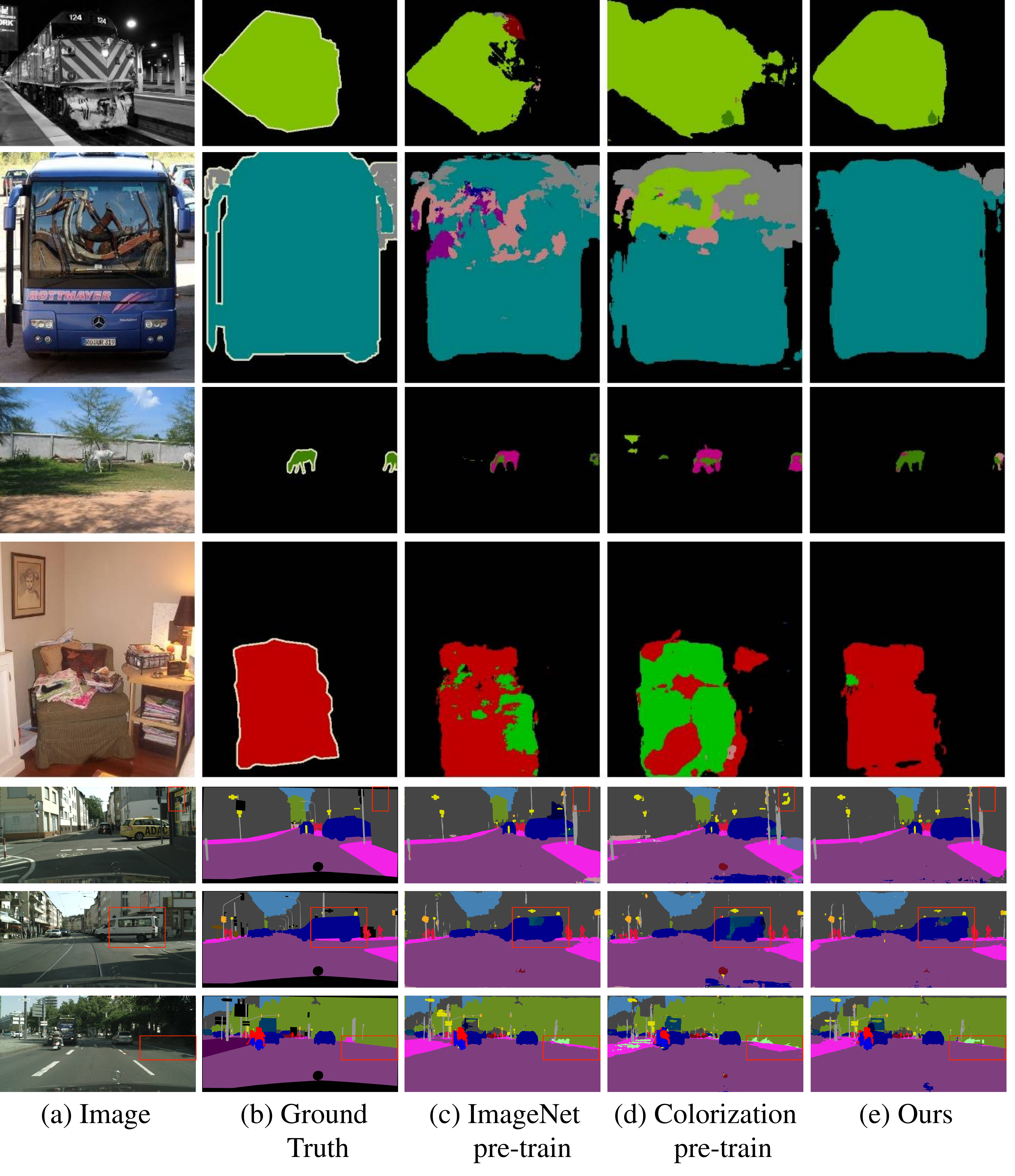}
    \caption{\small{Visual comparison on PASCAL VOC2012 validation set (top 4 rows) and CityScapes validation set (bottom 3 rows). (a) Image. (b) Ground Truth. (c) Results with ImageNet supervised pre-training. (d) Results with colorization pre-training. (e) Our results.\label{fig:vlz_main}}}
\end{figure}

\noindent
\textbf{Overall.}
Existing self-supervision works report segmentation results on PASCAL VOC2012 dataset.
The highest performance attained by existing self-supervision methods is learning by colorization~\cite{larsson2017colorization}, which achieves 38.4\% mIoU and 56.0\% mIoU with AlexNet and VGG-16 as the backbone network, respectively.
Therefore, we adopt learning by colorization as our proxy task here. 
With our M\&M tuning, we boost the performance to 42.8\% mIoU and 64.5\% mIoU with AlexNet and VGG-16 as the backbone network.
As shown in Table~\ref{tab:bench}, our method achieves state-of-the-art performance on semantic segmentation, outperforming~\cite{larsson2016learning} by 14.3\% and~\cite{larsson2017colorization} by 8.5\% when using VGG-16 as backbone network.
Notably, our M\&M self-supervision paradigm shows comparable results (0.3\% point of advantage) to its ImageNet pre-trained counterpart.
Furthermore, on PASCAL VOC2012 test set, our approach achieves 64.3\% mIoU, which is a record-breaking performance for self-supervision methods.
Qualitative results of this model are shown in Fig.~\ref{fig:vlz_main}.

We additionally perform an ablation study on the AlexNet setting.
As shown in Table~\ref{tab:bench}, with colorization task as pre-training, our class-wise connected graph outperforms `random triplets' by 1.9\%, suggesting the importance of class-wise connected graph.
With solving jigsaw-puzzles as pre-training task, our model performs even better than colorization pre-training.

\begin{table}[t]
    \centering
    \caption{\small{We test our model on PASCAL VOC2012 validation set, which is the generally accepted benchmark for semantic segmentation with self-supervised pre-training. Our method achieves the state-of-the-art with both VGG-16 and AlexNet architectures.\label{tab:bench}}}
    \scriptsize
      \begin{tabular}{l|lcc@{}}      
    \hline
    Method                                                   & Arch.                      & \begin{tabular}[c]{@{}c@{}}VOC12\\ \%mIoU.\end{tabular}     \\
    \hline
    ImageNet                                                 & VGG-16                     & 64.2               \\
    \hline
    Random                                                   & VGG-16                     & 35.0               \\
    Larsson \textit{et al.}~\cite{larsson2016learning}       & VGG-16                     & 50.2               \\
    Larsson \textit{et al.}~\cite{larsson2017colorization}   & VGG-16                     & 56.0               \\
    \hline
    Ours (M\&M + Graph, colorization pre-trained)            & VGG-16                     & \textbf{64.5}      \\
    \hline \hline        
    ImageNet                                                 & AlexNet                    & 48.0               \\
    \hline
    Random                                                   & AlexNet                    & 23.5               \\
    k-means~\cite{krahenbuhl2015data}                        & AlexNet                    & 32.6               \\
    Pathak \textit{et al.}~\cite{pathak2016context}          & AlexNet                    & 29.7               \\
    Donahue \textit{et al.}~\cite{donahue2016adversarial}    & AlexNet                    & 35.2               \\
    Zhang \textit{et al.}~\cite{zhang2016colorful}           & AlexNet                    & 35.6               \\
    Zhang \textit{et al.}~\cite{zhang2016split}              & AlexNet                    & 36.0               \\
    Noroozi et al.~\cite{noroozi2016unsupervised}            & AlexNet                    & 37.6               \\
    Larsson \textit{et al.}~\cite{larsson2017colorization}   & AlexNet                    & 38.4               \\
    \hline
    Ours (M\&M + Random Triplets, colorization pre-trained)  & AlexNet                    & 40.9               \\
    Ours (M\&M + Graph, colorization pre-trained)            & AlexNet                    & 42.8               \\
    Ours (M\&M + Graph, jigsaw-puzzles pre-trained)                & AlexNet                    & \textbf{44.5}      \\
    \hline
    \end{tabular}
\end{table}

\begin{table*}[ht]
\centering
\caption{\small{Per-class segmentation results on PASCAL VOC2012 val. The last row shows the additional results of our model combined with ImageNet pre-trained model by averaging their prediction probabilities. The results suggest the complementary nature of our self-supervised method with ImageNet pre-trained model.\label{tab:perclass}}}
\small
\resizebox{\linewidth}{!}{
	\begin{tabular}{l|cccccccccccccccccccc|c}
	    \hline
	             & aero & bike & bird & boat & bottle & bus & car & cat & chair & cow & table & dog & horse & mbike & person & plant & sheep & sofa & train & tv & mIoU. \\
	       \hline
	ImageNet     & 81.7 & \textbf{37.4} & \textbf{73.3} & 55.8 & 59.6 & 82.4 &  74.7 & 82.4 & 30.8  & 60.3 & 46.1 & \textbf{71.4} & 65.3 & \textbf{72.6} & 76.7 & \textbf{49.7} & 70.6 & 34.2 & 72.7 & \textbf{60.2} & 64.2 \\
	\hline 
	Colorization & 73.6 & 28.5 & 67.5 & 55.5 & 50.2 & 78.3 & 66.1 & 78.3 & 26.8 & 60.8 & \textbf{50.6} & 70.6 & 64.9 & 62.2 & 73.5 & 38.2 & 66.8 & \textbf{38.8} & 68.1 & 55.1 & 60.2 \\
	\hline 
	M\&M          & \textbf{83.1} & 37.0 & 69.6 & \textbf{56.1} & \textbf{62.9} & \textbf{84.4} & \textbf{76.4} & \textbf{82.8} & \textbf{33.4} & \textbf{61.5} & 44.7 & 67.3 & \textbf{68.5} & 68.0 & \textbf{78.5} & 42.2 & \textbf{72.7} & 37.2 & \textbf{75.7} &    58.6 & \textbf{64.5} \\
	\hline
	\hline 
	\begin{tabular}[c]{@{}l@{}}Ensemble\\ ImageNet+M\&M\end{tabular} & 84.5 & 39.4 & 76.3 & 60.3 & 64.6 & 85.4 & 77.7 & 84.1 & 35.6 & 63.6 & 50.4 & 70.6 & 72.0 & 73.6 & 80.1 & 50.2 & 73.7 & 37.6 & 77.8 & 66.6 & 67.4 \\
	\hline
	\end{tabular}
}
\end{table*}

\begin{figure*}[ht]
\centering
\includegraphics[width=\linewidth]{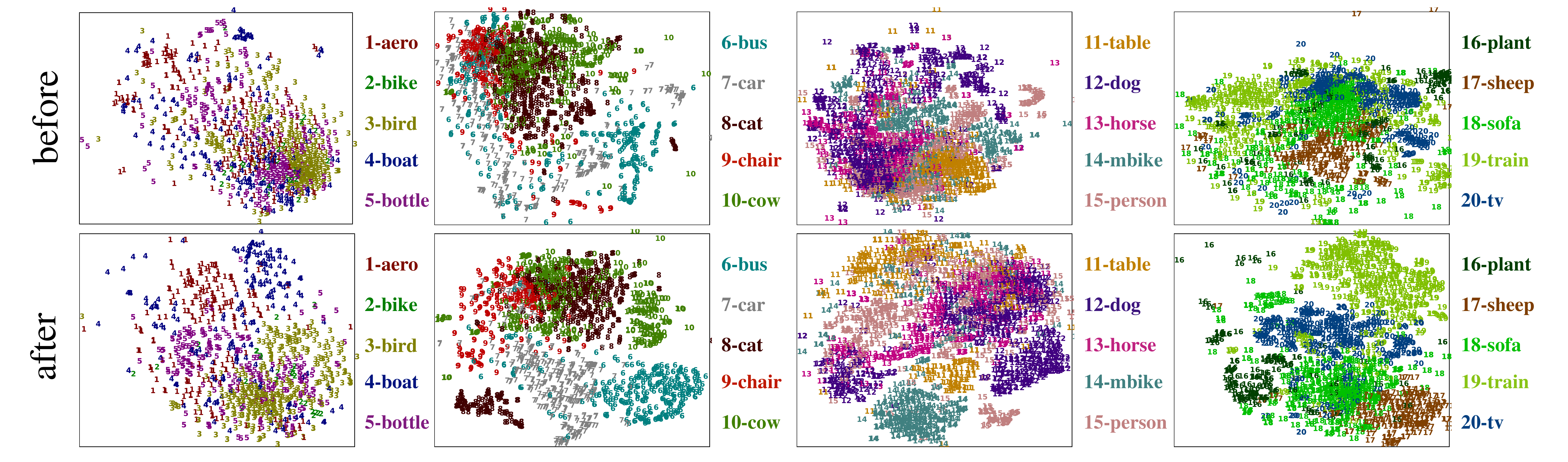}
\caption{\small{Feature distribution with and without the proposed mix-and-match (M\&M) tuning. We use 17,684 patches obtained from PASCAL VOC2012 validation set to extract features, and map the high-dimensional features to a 2-D space with t-SNE, along with their categories. For clarity, we split 20 classes into four parts in order. The first row shows the feature distribution of a naively fine-tuned model without M\&M, and the second row depicts the feature distribution of a model additionally tuned with M\&M. Note that the features are respectively extracted from the CNNs which have been fine-tuned to segmentation task, in this case, two CNNs have undergone the identical amount of data and labels. Best viewed in color.}}
\label{fig:tsne}
\end{figure*}

\noindent
\textbf{Per-class results.}
We analyze per-class results of M\&M tuning on PASCAL VOC2012 validation set. 
The results are summarized in Table~\ref{tab:perclass}. 
When compared our method with the baseline model that uses colorization\footnote{We obtain higher performance than reported with the released pre-training model of ~\cite{larsson2017colorization}.} as pre-training, our approach demonstrates significant improvements in classes including aeroplane, bike, bottle, bus, car, chair, motorbike, sheep, train.
A further attempt at combining our self-supervised model and the fully-supervised model (through averaging their predictions) leads to an even higher mIoU of 67.4\%.
The results suggest that self-supervision serves as a strong candidate complementary to the current fully-supervised paradigm.

\noindent
\textbf{Applicability to different proxy tasks.}
Besides colorization~\cite{larsson2017colorization}, we also explore the possibility of using Jigsaw Puzzles~\cite{noroozi2016unsupervised} as our proxy task.
Similarly, our M\&M tuning boosts the segmentation performance from 36.5\%\footnote{We use the released pre-training model of Jigsaw Puzzles~\cite{noroozi2016unsupervised} for fine-tuning and obtain a slightly lower baseline than the reported 37.6\% mIoU in the paper.} mIoU to 44.5\% mIoU.
The result suggests that the proposed approach is widely applicable to other self-supervision methods.
Our method can also be applied to randomly initialized cases. In PASCAL VOC 2012, M\&M tuning boosts the performance from 19.8\% mIoU to 43.6\% mIoU with AlexNet and from 35.0\% mIoU to 56.7\% mIoU with VGG-16.
The improvements of our method over different baselines are shown in Table~\ref{tab:baselines} for PASCAL VOC 2012.

\begin{table}[t]
    \caption{\small{The table shows the improvements of our method with different pre-training tasks. They respectively are Random (Xavier initialization) with AlexNet and VGG-16, Jigsaw Puzzles~\cite{noroozi2016unsupervised} with AlexNet and Colorization~\cite{larsson2017colorization} with AlexNet and VGG-16. Baselines are produced with naive fine-tuning. ImageNet pre-trained results are regarded as upper bound. Evaluations are conducted on PASCAL VOC2012 validation set and CityScapes validation set. Results on testing sets are shown in brackets.\label{tab:baselines}}}
  	\resizebox{.47\textwidth}{!}{
  		\centering
  		\small
	    \begin{tabular}{|l|c|c|c|c|c|@{}c@{}|c|c|}
	     \hline 
	     benchmark & \multicolumn{5}{c|}{PASCAL VOC2012} && \multicolumn{2}{c|}{CityScapes} \\
	     \hline
	     \hline
	     pre-train & Random & Jigsaw & Colorize & Random & Colorize && Random & Colorize\\ 
	     \hline
	     backbone & \multicolumn{3}{c|}{AlexNet} & \multicolumn{2}{c|}{VGG-16} && \multicolumn{2}{c|}{VGG-16}\\ 
	    \hline 
	    baseline & 19.8 & 36.5 & 38.4 & 35.0 & 60.2 && 42.5 & 57.5\\ 
	    \hline 
	    M\&M & 43.6 & \textbf{44.5} & 42.8 & 56.7 & \textbf{64.5 (64.3)} && 49.1 & \textbf{66.4 (65.6)}\\ 
	    \hline 
	    \hline
	    ImageNet & \multicolumn{3}{c|}{48.0} & \multicolumn{2}{c|}{64.2} && \multicolumn{2}{c|}{67.9}\\ 
	    \hline 
	    \end{tabular}
    }
\end{table}

\noindent
\textbf{Generalizability to CityScapes.}
We apply our method on CityScapes dataset. With colorization as pre-training, naive fine-tuning yields 57.5\% mIoU and M\&M tuning improves it to 66.4\% mIoU. The result is comparable with ImageNet pre-trained counterpart that yields 67.9\% mIoU.
With a random initialized network, M\&M could bring a large improvement from 42.5\% mIoU to 49.1\% mIoU.
The comparison can be found in Table~\ref{tab:baselines}.

\subsection{Further Analysis}

\noindent
\textbf{Learned representations.}
To illustrate the learned representations enabled by M\&M tuning, we visualize the sample distribution changes in the t-SNE embedding space.
As shown in Fig.~\ref{fig:tsne}, after M\&M tuning, samples from the same category tend to stay close while those from different categories are torn apart.
Notably, this effect is more pronounced on categories of aeroplane, bike, bottle, bus, car, chair, motorbike, sheep, train and tv, which aligns with the per-class performance improvements listed in Table~\ref{tab:perclass}.

\begin{figure}[t]
    \centering
    \includegraphics[width=0.9\linewidth]{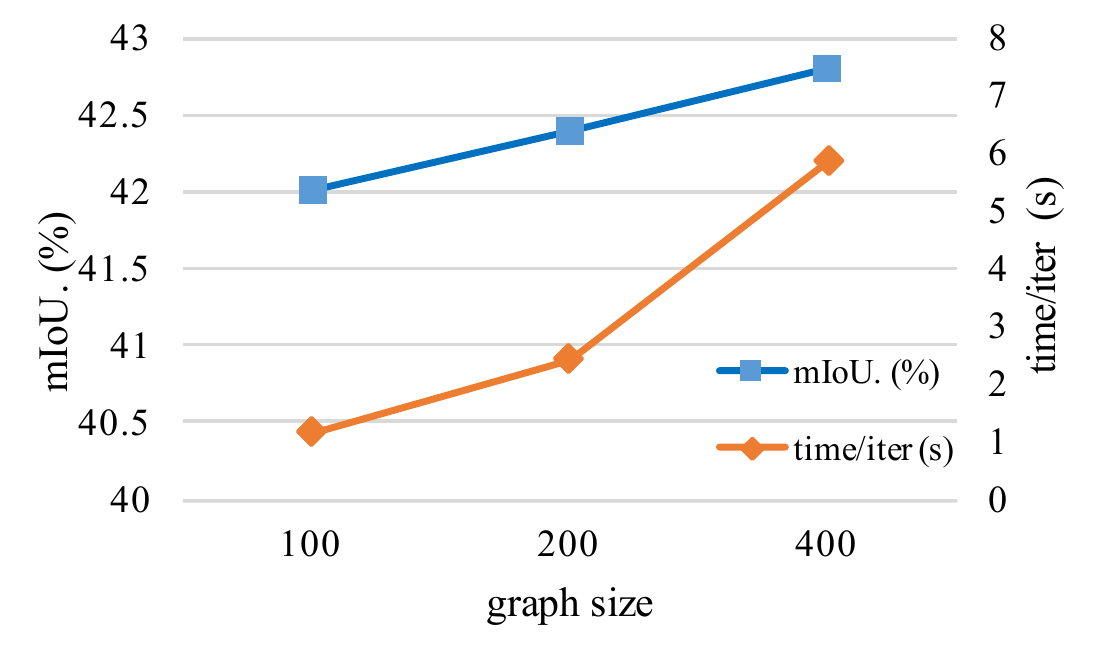}
    \caption{\small{The figure shows that a larger graph brings better performance, but costs a longer time in each iteration. We train the model with the same hyper-parameters for different settings and test on PASCAL VOC2012 validation set.\label{fig:analysis}}}
\end{figure}

\noindent
\textbf{The Effect of graph size.}
Here we investigate how the self-supervision performance is influenced by the graph size (the number of nodes in the graph), which defines the number of triplets that can be discovered.
Specifically, we set the image batch size to be $\{10, 20, 40\}$, so that the number of nodes is $\{100, 200, 400\}$, as shown in Fig.~\ref{fig:analysis}.
The comparative study is performed on AlexNet with learning by colorization~\cite{larsson2017colorization} as initialization. We have the following observations.
On the one hand, a larger graph leads to a higher performance, since it brings more diverse samples for more accurate metric learning.
On the other hand, a larger graph takes longer time for processing, since a larger batch size of images is fed in each iteration.

\noindent
\textbf{Efficiency.}
The previous study suggests that performance and speed trade-off can be enabled through graph size adjustment.
Nevertheless, our graph training process is very efficient. It costs respectively $3.5$ hours and $5.8$ hours on a single TITAN-X for AlexNet and VGG-16, which are much faster than conventional ImageNet pre-training or any other self-supervised pre-training task.

\begin{figure}[!ht]
    \centering
    \includegraphics[width=\linewidth]{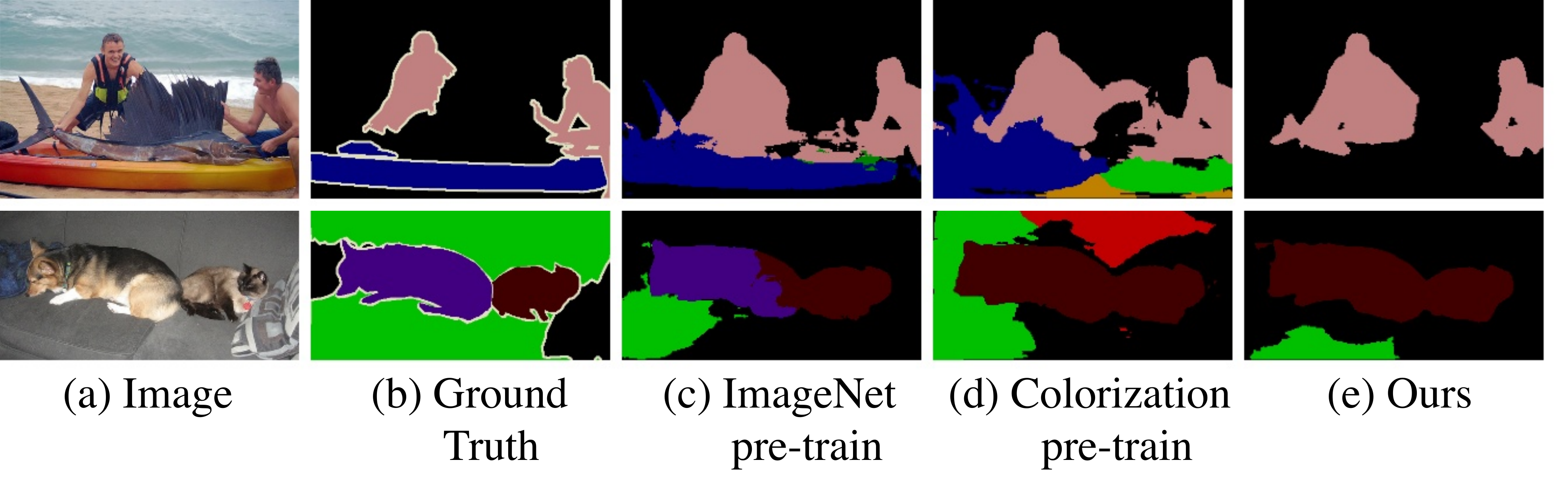}
    \caption{\small{Our failure cases. (a) Image. (b) Ground Truth. (c) Results with ImageNet supervised pre-training. (d) Results with colorization pre-training. (e) Our results.\label{fig:vlz_fail}}}
\end{figure}

\noindent
\textbf{Failure cases.}
We also include some failure cases of our method, as shown in Fig.~\ref{fig:vlz_fail}.
The failed examples can be explained as follows.
Firstly, patches sampled from thin objects may fail to reflect the key characteristics of the object due to the clutter, so the boat in the figure ends as a false negative.
Secondly, our M\&M tuning method inherits its base model (i.e., colorization model) to some extent, which accounts for the case in the figure that the dog is falsely classified as a cat.

\section{Conclusion}
\label{sec:conclusion}

We have presented a novel `mix-and-match' (M\&M) tuning method for improving the performance of self-supervised learning on semantic image segmentation task.
Our approach effectively exploits mixed image patches to form a class-wise connected graph, from which triplets can be sampled to compute a discriminative loss for M\&M tuning.
Our approach not only improves the performance of self-supervised semantic segmentation with different proxy tasks and different backbone CNNs on different benchmarks, achieving state-of-the-art results, but also outperforms its ImageNet pre-trained counterpart for the first time in the literature, shedding light on the enormous potential of self-supervised learning.
M\&M tuning is potentially to be applied to various tasks and worth further exploration. Future work will focus on the essence and advantages of multi-step optimization like M\&M tuning.

\noindent
\textbf{Acknowledgement}. This work is supported by SenseTime Group Limited and the General Research Fund sponsored by the Research Grants Council of the Hong Kong SAR (CUHK 14241716, 14224316. 14209217). 

\small{
\bibliography{egbib}
\bibliographystyle{aaai}
}
\end{document}